\documentclass[letterpaper]{article}
\usepackage{preprint-layout}
\usepackage[hyphens]{url}
\usepackage{graphicx,booktabs,caption,placeins}
\usepackage{natbib}
\usepackage[hidelinks]{hyperref}
\usepackage{orcidlink}

\hypersetup{
  pdftitle={Detecting Missing Expected Events in Temporal Grounding with Opposite-Action Videos},
  pdfauthor={Bibhor Puhan, Sanchita Kiran, Satwik Sai Prakash Sahoo, Shubhankar Mahanta}
}

\setcitestyle{aysep={}}
\let\cite\citep

\newcommand{\paperauthor}[4]{%
  \mbox{\href{https://openreview.net/profile?id=~#2}{#1}%
  \textsuperscript{#3}\kern.12em\raisebox{.35ex}{\orcidlink{#4}}}%
}

\title{Detecting Missing Expected Events in Temporal Grounding with Opposite-Action Videos}
\author{%
  \paperauthor{Bibhor Puhan}{Bibhor_Puhan2}{1}{0009-0007-0562-7258},\enspace
  \paperauthor{Sanchita Kiran}{Sanchita_Kiran1}{2}{0009-0009-1916-2038},\enspace
  \paperauthor{Satwik Sai Prakash Sahoo}{Satwik_Sai_Prakash_Sahoo1}{3}{0009-0002-1943-9323},\enspace
  \paperauthor{Shubhankar Mahanta}{Shubhankar_Mahanta1}{1}{0009-0006-7314-6627}%
}
\affiliations{%
  \textsuperscript{1}\,Siksha `O' Anusandhan, Bhubaneswar, India\\
  \textsuperscript{2}\,Silicon University, Bhubaneswar, India\\
  \textsuperscript{3}\,Odisha University of Technology and Research, Bhubaneswar, India%
}

\begin{document}

\maketitle

\begin{abstract}
A video-language grounder should be able to detect both when an event occurs and when it does not occur, which is an omission anomaly. However,
previous tests involve either unrelated videos or language model edited sentences. The latter can be identified based
on their content. We maintain the original sentences, and pair them with videos annotated to indicate that the event
does not occur. These may show the opposite event occurring. Annotations were verified by human raters on a sample. On
a held out test set, six video-language grounders from three model families flag videos showing opposite events less frequently (by
26.8 to 43.4 percentage points) than unrelated videos. When a grounder is trained to reject altered sentences, it
performs better than the baseline model on that altered sentence benchmark, but not on the opposite-action videos. A
previously existing training-free test, which asks the same yes/no question with frames in order and reversed, can
catch many of these missed examples at the cost of approximately 5 additional false alarms per 100 original videos.
Real videos showing opposite events should be included in tests of detecting event absence.
\end{abstract}

\section{Introduction}
\label{sec:intro}

\begin{figure}[t]
\centering
\includegraphics[width=\columnwidth]{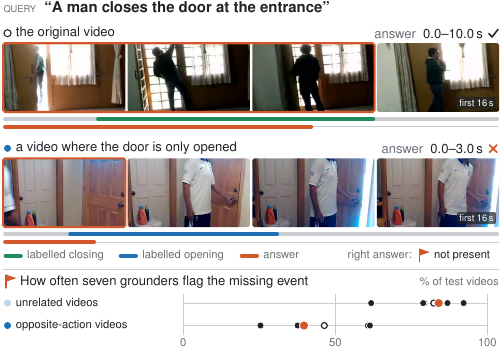}
\caption{Qwen3-VL gives a time span for the sentence on its original video and on a video where the door is only
opened. Bars: annotated query action (green), annotated opposite action (blue), answer (orange). Below: detection
rates of seven grounders on the test set (open circle: RaTSG, descriptive).}
\label{fig:teaser}
\end{figure}
In many video monitoring tasks, it is necessary to detect both whether and when an event occurred. A care system may
need to know when a patient stood up, and a production line when a component was screwed in. When asked such
questions, video-language models should provide the time the event occurs, or flag the event as missing if it does not
occur. Wrongly associating a time with an event that did not occur is a costly mistake. It is particularly problematic
when a similar but incorrect event occurs. For example, the model may be asked when a man closes the door at the
entrance, in a video where he only opens it. Any time span it gives is then wrong, and the right answer is to flag the
closing as missing.

From an anomaly-detection perspective, such a missing event is an omission~\cite{liu2021eventoutlier}, judged against
the sentence as its context~\cite{chandola2009anomaly}. Grounding benchmarks mostly build missing events from unrelated
videos or from sentences edited by a language model (Section~\ref{sec:related}). Unrelated videos are easy to flag.
Edited sentences have a different weakness: they can often be detected from the text alone. On
RA-RFT's~\cite{lee2025learningrefuse} publicly available \mbox{HI-Charades} test file, which contains 1,097 edited
sentences, a classifier that reads only the text, without the video, trained and tested in five folds split by video
on the 1,088 edited sentences with video and their originals, told edited from unedited sentences with an F1 of 77.1
in our evaluation. This was higher than the 71.8 that RA-RFT achieved with the video. To our knowledge, no
grounding benchmark keeps the sentence fixed and shows a real video in which the opposite action occurs.

We test this setting: the human-written sentence is kept fixed and only the video changes. Besides its original video,
each sentence is paired with up to three videos in which, according to the Charades annotations, the event never
happens: one unrelated to the query object, one with other actions on the same object and one showing only the
opposite action. An example is shown in Figure~\ref{fig:teaser}. As annotations are not perfect, we had people watch a
sample of these videos. They were not told what type of video they were looking at. They rated 92 out of 98
opposite-action videos as not containing the event.

Each model was scored by its false alarms on the original videos and by its detection rate on each type of video
without the event, that is, the share of those videos it flags. On the test set, all six video-language grounders
flagged opposite-action videos far less often than unrelated ones, with gaps of 26.8 to 43.4 points. For the four
grounders checked on the watched videos, the gap stayed large, and for three of them it was at least as large as with
the labels alone. A gap also appears on Charades-Ego~\cite{sigurdsson2018charadesego}, which films the same kind of household actions from the actor's own view, although there it rests on the annotations alone. RaTSG, which has
no language model, showed a similar gap of 37.6 points (descriptive). RA-RFT's refusal training did not carry over to
opposite-action videos: RA-RFT flagged 14.5 points more of its own edited sentences than its untrained base model,
Qwen2.5-VL-7B, but only 0.7 points more of our opposite-action videos.

A simple check that needs no training helped more. We asked the same model the same yes/no question twice, once with
the video frames in order and once in reverse. The difference between the two answers was used as the score,
following MotionCD~\cite{kong2025mhbench}. With a threshold set on the development set, this check identified many of
the opposite-action videos that the grounding models missed, at a cost of 3.4 to 5.1 extra false alarms per 100
original videos.

The paper makes three contributions:
\begin{itemize}
  \item We build a test that keeps a human-written sentence fixed and changes only the video, to real videos whose
        annotations show that the event is absent (Section~\ref{sec:protocol}).
  \item We show, on a test set fixed in advance, that opposite-action videos are the hardest case for all seven
        grounders (Section~\ref{sec:results}).
  \item We find that RA-RFT's refusal training does not carry over to opposite-action videos, and that a known
        training-free reversed-frames check recovers many of them, more than shuffled frames for two of the three
        model families (Section~\ref{sec:results}).
\end{itemize}

\section{Related Work}
\label{sec:related}
\paragraph{Contextual Anomalies and Omission Detection.}
A contextual anomaly is an observation that is anomalous with respect to its
context~\cite{chandola2009anomaly,ruff2021unifying}. An omission is a specific type of anomaly in which an expected
event does not occur in an observed sequence of events~\cite{liu2021eventoutlier}. In video anomaly detection,
AnyAnomaly and LaGoVAD use text descriptions as context for detecting anomalies in videos, but in both cases the
anomalies are always present within the video~\cite{ahn2026anyanomaly,liu2026lagovad}. VALU includes a multiple-choice
task where the model is presented with descriptions of anomalies, generated by a language model, that do not happen in
the video, and must recognise that they do not happen~\cite{he2026valu}. In procedure monitoring, missing steps are
explicitly evaluated. IndustReal counts a step reported as performed but never completed as a false
positive~\cite{schoonbeek2024industreal}. \citet{seminara2024taskgraph} note that falsely detecting a step as
performed hides the error.

\paragraph{Missing Events and Near Negatives.}
Previous grounding benchmarks create missing events by either using videos unrelated to the
query~\cite{dong2024ratsg,flanagan2025untruth}, modifying the query to no longer match the video using a language
model~\cite{lee2025learningrefuse,anymoment2027,cedvs2027} or obfuscating the visual evidence of the correct
answer~\cite{accuracynotevidence2027}. HRVTG~\cite{yi2026hrvtg} retains modified queries only when its own grounder
predicts the event to be missing. In previous grounding work the answer that the event is missing is referred to as a
refusal, but we will use the term flag. In out-of-distribution and anomaly detection work, far and near negatives are
distinguished. Near negatives closely resemble the expected class~\cite{yang2022openood,vaze2022openset}. These are
harder to detect than far negatives~\cite{zhang2024openood15}. Training using examples of outliers (outlier exposure)
does not always help with near outliers~\cite{zhang2023mixoe}. If synthetic anomalies do not look like real anomalies,
they are not useful~\cite{yoo2023augmentation}. Refusal training can therefore be thought of as outlier exposure but
using negatives created by a language model. Query modifications can already reverse the action described while the
video stays the same~\cite{anymoment2027,cedvs2027}. We have created negatives by keeping the original human-written
query and matching it to different real videos. The annotations of these videos show that the event being queried is
missing. We have ordered our negatives from far to near negatives. The nearest negatives are videos that show the
opposite action.

\paragraph{Time Order as a Signal.}
In video anomaly detection, the direction of time is used as a training task~\cite{georgescu2021ssmtl}, and
transformation-based detectors score an input by how a model's output changes under a
transform~\cite{golan2018geometric}. TimeCatch~\cite{hradil2026timecatch} already casts temporal grounding as anomaly
detection, using videos with swapped frames. Reversed frames are also used with video-language models.
MotionCD~\cite{kong2025mhbench} compares a model's yes/no answer on a video and on its reversed copy to correct that
answer, and Caved or Convinced~\cite{cao2026caved} compares forward and reversed frames to decide whether to answer.
ArrowGEV~\cite{yu2026arrowgev} trains a grounder on reversed copies but gives it no way to flag a missing event, and
TAI~\cite{shin2026before} adds the forward-minus-reversed difference inside the model. Opposite-action pairs have also
been mined from Charades for retrieval~\cite{bagad2025chirality}. We instead use the reversed-frames difference to flag
grounding answers, at a fixed false-alarm budget.

\begin{figure*}[!t]
\centering
\includegraphics[width=\textwidth]{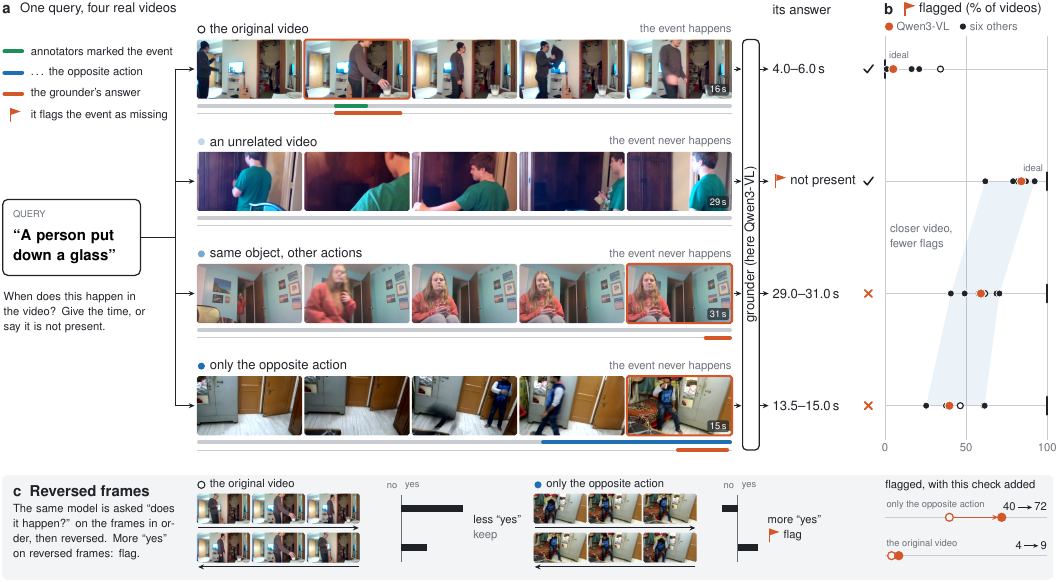}
\caption{(a) One sentence on its original video and three videos without the event, with Qwen3-VL's answers;
bars as in Figure~\ref{fig:teaser}. (b) Detection rates of seven grounders on the test set; the top row is false
alarms on 450 original videos (open circle: RaTSG, descriptive). (c) The reversed-frames check on the same example and
its effect for Qwen3-VL on the 295 paired sentences, with the threshold fixed on the development set.}
\label{fig:main}
\end{figure*}

\section{Test and Protocol}
\label{sec:protocol}

\subsection{One Query, Four Videos}
\label{sec:test}

Every evaluation item contains a sentence and a video. The model must find when the event happens in the video or
determine that it is not present. Sentences and their time spans are taken from Charades-TimeLens~\cite{zhang2025timelens},
a re-annotation of Charades-STA~\cite{gao2017tall}. The videos are taken from the Charades test
split~\cite{sigurdsson2016hollywood}. Charades contains temporal annotations for 157 action classes, scripts and
free-text descriptions. These are used to identify videos where the event is not present.

\paragraph{Query Action.}
Each sentence is mapped to action classes in Charades by matching verbs and object names to static lists of words.
Only sentences that have exactly one action class annotated at a time overlapping the time span of the sentence are
used. This action class becomes the query action. If more than one action is mentioned within the sentence for the
same object or the sentence refers to dressing (putting on and taking off clothing) then the sentence is discarded.

\paragraph{No Event.}
The event is considered not present in a video if none of the action classes matched to the sentence are annotated in
it, and none are mentioned in its script or descriptions. Each sentence is paired with its original video and up to
three videos that meet this criterion. As the sentence is the same for all of them, the presence or absence of the
event cannot be determined from the text alone.

\paragraph{Not Present Videos.}
For each sentence, up to one valid video is selected from each of three categories using a fixed random seed (Figure~\ref{fig:main}a).
\begin{itemize}
  \item \emph{Unrelated:} the video has no annotated action involving the query object.
  \item \emph{Same Object:} the video has other actions involving the object and is from the same room category. No
        opposite action is annotated.
  \item \emph{Opposite Action:} the video has an annotated opposite action. For example, if the query is to open a
        door, the video will have an annotation for closing a door. Where possible, the video is from the same room
        category.
\end{itemize}
The opposite actions are defined using 27 pairs of action classes, such as opening/closing, putting/taking, on/off and
sitting/standing. The videos therefore increase in similarity to the query: first videos with
no interaction with the query object, then videos with other interactions with it and finally videos with the
opposite action. Opposite actions often share the object and a similar motion with the query.

\paragraph{Development Set.}
There are 450 sentences in the development set, with up to 30 sentences for each query action. Priority is given to
sentences that have a valid opposite-action video, and the rest are filled with those that have a same-object video.
The development set contains 1,685 items: 450 original videos, 450 unrelated videos, 335 same-object videos and 450
opposite-action videos.

\paragraph{Test Set.}
The test set contains no sentences and no sentence and video pairs from the development set. It was fixed before any
model was evaluated on it. After excluding 14 same-object videos where the query action is holding an object, 1,537
of its 1,551 items remain: 450 original videos, 450 unrelated videos, 342 same-object videos and 295 opposite-action
videos. In these videos the object is taken, put, thrown or tidied, which implies that it is held even though holding
is not annotated. Videos, but not sentences, may recur across the two sets.

Section~\ref{sec:viewing} reports a video audit of how often the annotations miss an event.

\subsection{Checking the Labels by Eye}
\label{sec:viewing}

There may be cases where the annotations miss the query action although it happens in the video. We therefore
verified a sample by watching it. Using a fixed random seed, 300 items were taken, with no more than one item per sentence: 210 from the test
set (100 opposite-action, 50 same-object, 30 unrelated and 30 original videos) and 90 absent-event probes released by
HRVTG~\cite{yi2026hrvtg}. Three raters watched 200 items each, so that
each item was judged twice independently. Raters were not informed about which category an item fell into. The raters
simply returned a `yes', `no' or `unsure' response, and a justification. They were asked whether the query action
happened to its object. If responses differed, a third rater who had not seen the item gave an answer. Where possible,
majority voting was used.

\paragraph{Results.}
In total, 100 opposite-action videos were watched and 98 received a verdict. Of these, 92 were judged not to contain
the queried event (93.9\%, Wilson interval $[87.3, 97.2]$), higher than the threshold of 90\% fixed beforehand
(planned). Two videos could not be decided. All 30 unrelated videos and 47 of 50 (94\%) same-object videos were judged
not to contain the event, and all 30 original videos were judged to contain it. Agreement between pairs of raters
ranged from 85\% to 93\% ($\kappa$: 0.50--0.80).

HRVTG keeps only probes that its own grounder deems not to contain the event. Of the 60 probes whose own-video
annotations contradict them, 24 (40.0\%, Wilson interval $[28.6, 52.6]$; planned) were identified as containing the
event by the raters, compared to 3 of the remaining 30. If a model filters its own negatives, examples where the event
does occur can still be retained.

\subsection{Flags and False Alarms}
\label{sec:scoring}

A temporal grounder returns the time interval of the event that a sentence describes. We require it to either give
this interval or flag the event as not present. Grounders based on Qwen2.5-VL are prompted with the RA-RFT
prompt~\cite{lee2025learningrefuse}, which asks for the time interval if a relevant segment exists in the video and
otherwise for a statement that the video does not contain one. Qwen3-VL is prompted with two sentences asking for the
time interval or ``not present''. PerceptionLM answers in its own format, frame numbers: ``Does `[sentence]' happen in
this 32 frame video? If it does not happen, answer `not present'. If it happens, give its start and end frame in the
format [start, end], where start and end are frame numbers between 0 and 31.'' Frames $a$ to $b$ of a video of length
$L$ seconds become the interval from $aL/32$ to $(b+1)L/32$.

\paragraph{Flagging the Model Output.}
We parse the model outputs with a fixed parser, written before the first run. For the RA-RFT prompt, an answer with no
time interval is flagged. If the answer includes a time interval and extra text, the correction field decides: if the
field is \texttt{NIL} the answer is not flagged, otherwise it is flagged. For the two-sentence prompt and for
PerceptionLM, an answer is flagged only if it has no time interval and states that the event is not present (or uses
similar wording). Under every prompt, an answer that cannot be read is never flagged. On the test set there were 9
unreadable outputs from the untrained Qwen2.5-VL and none from the other Qwen-based grounders.
PerceptionLM's bare ``n'' answers are treated in Table~\ref{tab:main}.

\paragraph{False Alarms and Detection Rate.}
If the model flags the original video of a sentence, it is a false alarm. If the model flags a video where the event
does not occur, it is a detection. We report the detection rate separately for unrelated, same-object and
opposite-action videos. We consider unrelated videos to be far out-of-distribution examples and the other two
categories to be near out-of-distribution examples~\cite{yang2022openood}. The main comparison is the gap between the
detection rate on unrelated videos and that on opposite-action videos. This is calculated on the 295 test sentences
that have both video types. The 95\% confidence interval is estimated from 2,000 bootstrap samples of the original
videos. The sentences of each video are kept together, as several sentences can belong to one video.

\paragraph{Checking on Reversed Frames.}
We also ask the grounder whether the action occurs in the video, with the same frame sampling as the grounding. The
question is ``Does this video clip show the following action: \{action\}? Answer yes or no.'', where the action is the
Charades class label of the query action. The score is calculated as:
\begin{equation}
  m = \log \sum_{t \in \{\mathrm{Yes}, \mathrm{yes}\}} e^{\ell_t} - \log \sum_{t \in \{\mathrm{No}, \mathrm{no}\}} e^{\ell_t}{,}
\end{equation}
where $\ell_t$ is the logit for the token $t$ at the first answer position. The question is repeated with the frames
in reversed order, and the check uses the difference $m_{\mathrm{reversed}} - m_{\mathrm{forward}}$. This is because
the opposite action may look like the action we are asking about when it is played backwards.

MotionCD~\cite{kong2025mhbench} uses this difference to correct yes/no predictions by subtracting a scaled version of
it from the forward score. We instead threshold the difference. We flag an item if it is flagged by the grounder or
the difference is above the threshold. The threshold was chosen on the development set to add false alarms on at most
5\% of the original videos. This threshold was then fixed for testing. As a control, we replace the reversed frames
with the same frames in a random order, drawn with a fixed seed for each item.

\subsection{Models and Settings}
\label{sec:models}

Seven temporal grounders were tested. Six use vision-language models from three families and one uses pre-extracted
features. A further model, Molmo2-O-7B, did not pass our grounding quality check in a pilot and is not
reported. Four models use Qwen2.5-VL-7B as their base model, allowing comparison of different training methods:
the instruction fine-tuned model itself without additional training (Qwen2.5-VL-7B-Instruct~\cite{bai2025qwen25vl});
Time-R1-7B~\cite{wang2025timer1}, fine-tuned for temporal grounding using reinforcement learning;
ArrowGEV-7B~\cite{yu2026arrowgev}, fine-tuned using reinforcement learning with a reward that combined grounding
accuracy on the original video with a comparison of the model's predictions on the original and the reversed video;
and RA-RFT~\cite{lee2025learningrefuse}, the released checkpoint fine-tuned from Time-R1 to refuse queries that a
language model had edited so that they no longer match the video. Qwen3-VL-8B-Instruct~\cite{bai2025qwen3vl} and
PerceptionLM-8B~\cite{cho2025perceptionlm} are two further families, used as released; PerceptionLM answers in its
native format of frame numbers.
RaTSG~\cite{dong2024ratsg} is a feature-based model. It uses pre-extracted video features. It was trained on
Charades-STA with negative examples being unrelated query and video pairs. The model and checkpoint provided by the
authors were used to reproduce the published results. Its results are reported descriptively (Table~\ref{tab:main},
footnote d).

\paragraph{Settings and Cost.}
The Qwen models used 2 frames per second for the video input. The maximum number of pixels per video was 2,809,856
and the maximum number of new tokens was 512. The PerceptionLM model had its own settings with 32 frames at
448$\times$448 pixels and 32 new tokens. All of the vision-language grounding models used greedy decoding in the
bfloat16 format and were run on a single NVIDIA RTX PRO 6000 Blackwell GPU. Each model was run once. The reversed
frame check took around 1.4 seconds per item to run, measured on 128 test items. This was similar to one grounding
from ArrowGEV (1.48\,s) and around 2.6 times as long as one response from Qwen3-VL (0.53\,s).

\section{Results}
\label{sec:results}

\subsection{Opposite actions are missed}

Among absent-event videos, all seven grounders are most likely to flag an unrelated video, less likely to flag a same-object video and least likely to flag an opposite-action video (Figure~\ref{fig:main}b).
The paired gap ranges from 26.8 to 43.4 points for the six video-language grounders (RaTSG, 37.6 points, descriptive).
The development set showed similar gaps before the test set existed (19.6--38.2 points).
Previous open-set studies also found near negatives harder to detect using AUROC~\cite{yang2022openood,vaze2022openset}.
Here we report flag rates.

\begin{table*}[!t]
\centering
\small
\setlength{\tabcolsep}{2pt}
\begin{tabular*}{\textwidth}{@{}l@{\extracolsep{\fill}}rrrrrrrr@{}}
\toprule
 & \multicolumn{4}{c}{Flag rate (\%)} & \multicolumn{1}{c}{Originals (\%)} & \multicolumn{1}{c}{295 pairs} & \multicolumn{2}{c}{Viewing analysis: gap (points)\textsuperscript{e}} \\
\cmidrule(lr){2-5}\cmidrule(lr){6-6}\cmidrule(lr){7-7}\cmidrule(l){8-9}
Grounder or prompt & Orig. & Unrel. & Same obj. & Opp. & $\mathrm{IoU}{\geq}0.3$ & Gap [95\% CI] & Confirmed [95\% CI] & Annotation \\
\midrule
Qwen2.5-VL-7B (post hoc) & 21.1 & 86.9 & 68.7 & 60.7 & 35.3 & $+26.8\,[+19.9,+33.8]$ & --- & --- \\
Time-R1               & 16.4 & 79.3 & 57.6 & 40.0 & 43.1 & $+38.3\,[+31.5,+44.9]$ & $+42.4\,[+31.2,+52.7]$ & $+39.3$ \\
ArrowGEV              &  6.2 & 61.8 & 40.6 & 25.4 & 49.1 & $+34.9\,[+27.4,+42.3]$ & $+35.7\,[+24.9,+45.7]$ & $+36.4$ \\
RA-RFT                & 16.4 & 92.2 & 70.5 & 61.4 & 44.9 & $+30.8\,[+24.4,+37.4]$ & $+33.5\,[+22.6,+44.1]$ & $+30.9$ \\
Qwen3-VL-8B (two-sentence)\textsuperscript{a} & 5.1 & 84.0 & 59.1 & 39.7 & 57.8 & $+43.4\,[+36.4,+50.0]$ & --- & --- \\
Qwen3-VL-8B (RA-RFT)\textsuperscript{b} & 24.9 & 97.1 & 82.7 & 77.3 & 18.7 & $+19.0\,[+14.3,+23.8]$ & $+24.3\,[+15.3,+34.2]$ & $+19.8$ \\
PerceptionLM-8B\textsuperscript{c} & 1.1 & 78.9 & 49.1 & 37.6 & 62.7 & $+39.3\,[+32.1,+46.0]$ & --- & --- \\
\quad Runner-up prompt\textsuperscript{c} & 0.7 & 56.7 & 29.2 & 23.7 & 64.0 & $+34.2\,[+26.6,+41.5]$ & --- & --- \\
RaTSG (descriptive)\textsuperscript{d} & 34.2 & 82.4 & 61.4 & 46.4 & 22.9 & $+37.6\,[+30.8,+44.4]$ & --- & --- \\
\bottomrule
\end{tabular*}
\caption{Charades test set. Flags on original videos are false alarms. The paired gap is the unrelated minus opposite-action detection rate over 295 paired sentences, in percentage points. All rows are planned unless marked otherwise.}
\label{tab:main}
\par\smallskip\small\raggedright
\textsuperscript{a}Main row; chosen by a rule fixed before the test run, after RA-RFT's prompt gave empty intervals for
156 of 332 answers on original development videos.
\textsuperscript{b}142 of 338 answers on original videos are empty intervals, counted as answers.
\textsuperscript{c}Prompt chosen on the development set after a first prompt failed its planned test (gap $+9.2$); 123
bare ``n'' answers count as unreadable (as flags: gap $+35.6$).
\textsuperscript{d}Trained on longer intervals (7\,s against 2--3\,s here).
\textsuperscript{e}All 450 unrelated videos against the opposite-action videos raters confirmed empty
(Section~\ref{sec:viewing}), beside the same gap with all 295 opposite-action videos by annotation.
\end{table*}

Whether an empty interval counted as an answer or a flag, the gap persisted for all four prespecified grounders on both development and test data (planned).
Both prompts yield this gap for Qwen3-VL (Table~\ref{tab:main}).
Changing a single sentence in PerceptionLM's prompt increases its detection of unrelated videos from 13.6\% to 78.9\%, with little change in grounding performance (63.8\% versus 62.7\%, $\mathrm{IoU}\geq0.3$, post hoc).
For PerceptionLM's three prompts (Table~\ref{tab:main}, footnote c), the pairwise confidence intervals for differences in gaps on the normal $z$ scale
include zero (post hoc): the prompt moves the flagging threshold, not the gap.

All five Qwen grounders show this gap for both sentence lengths.
For sentences of six words or fewer the gap is 13.4 to 29.6 points larger, with all but one confidence interval above zero (post hoc).
Other discrepancies may trigger a flag, so the opposite-action flag rate gives an upper bound on detection of direction alone.

The four planned annotation-based gaps on Charades-Ego range from 19.3 to 38.3 points, each with a lower confidence bound above 10.
One rater found the query action in \mbox{11 of 38} opposite-action videos.
On Charades-Ego, flag rates were similar for same-object and opposite-action videos (post hoc).

The planned test on watched videos holds for all four grounders: the gaps remain positive when only opposite-action videos with rater-confirmed absences are considered, with three exceeding the corresponding annotation-based estimate.
ArrowGEV's gap changes by less than one point (Table~\ref{tab:main}).

\subsection{Gains from refusal training}
\label{sec:refusal-training}

\begin{figure}[!htb]
\centering
\includegraphics[width=0.91\columnwidth]{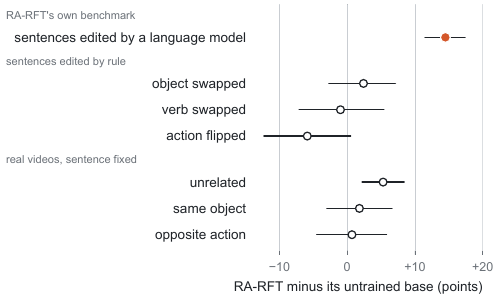}
\caption{Difference in flag rate between RA-RFT and Qwen2.5-VL-7B, in percentage points, with 95\% confidence
intervals. The rule-made edits span the development and test sets. A rater identified the flipped action in 5 of
40 original videos (12.5\%).}
\label{fig:training}
\end{figure}

RA-RFT is trained to flag sentences edited by a language model~\cite{lee2025learningrefuse}, which makes it outlier
exposure~\cite{hendrycks2019oe} with outliers generated by a language model (Section~\ref{sec:related}). We compare
RA-RFT with both its untrained base (Qwen2.5-VL-7B) and its starting point for refusal training (Time-R1). RA-RFT,
Qwen2.5-VL-7B and Time-R1 all use RA-RFT's grounding prompt. For the 1,088 edited HI-Charades sentences for which there
is video data, RA-RFT has improved detection of edited sentences by 14.5 percentage points $[+11.5, +17.4]$ compared with
the base (planned). It has increased detection of opposite-action videos by 0.7 percentage points $[-4.5, +5.8]$. The
gain on edited sentences is therefore 13.8 percentage points larger $[+8.0, +19.7]$ than on opposite-action videos
(Figure~\ref{fig:training}). Outlier exposure often gains little on near outliers~\cite{zhang2023mixoe,ding2022dra}.
In video-language tasks, models trained to detect one type of negative data gain most on that type of
data~\cite{bansal2023videocon,hsieh2023sugarcrepe}.

Compared to Time-R1, RA-RFT detects 23.5 percentage points more edited sentences and 21.4 percentage points more
opposite-action videos. The difference between the improvements in detection of edited sentences and opposite-action
videos is small, at just 2.2 percentage points $[-3.6, +7.9]$. This means that RA-RFT recovers the drop in detection of
opposite-action videos seen after grounding training. Both Time-R1 and ArrowGEV detect fewer videos of all types than the
base with this prompt (post hoc). This agrees with previous grounding
results~\cite{accuracynotevidence2027,anymoment2027}.

Relative to the base, RA-RFT has a lower false-alarm rate on original videos (16.4\% vs.\ 21.1\%), and returns
intervals with an $\mathrm{IoU}\geq0.3$ more frequently (44.9\% vs.\ 35.3\%). On Charades-Ego, by annotations, it
detects 5--11 percentage points more absent-event videos than the base (descriptive).

The yes/no answers produced by the base and fine-tuned Qwen2.5-VL checkpoints agree on 96.2\%--98.3\% of examples in the
development, test and Charades-Ego sets (planned). Therefore, the checkpoints differ in when they decide to flag, not in
what they see.

RA-RFT's gap between unrelated and opposite-action videos is smaller than Time-R1's in percentage points (30.8 vs.\
38.3), with RA-RFT near the ceiling on unrelated videos (92.2\%). Converting detection rates to $z$-scores, the
difference between detection rates in unrelated and opposite-action videos is $1.13\;[0.89, 1.42]$ for RA-RFT, and
$1.04\;[0.84, 1.24]$ for Time-R1 (post hoc). Therefore, RA-RFT's smaller gap reflects detection near the ceiling, not a
smaller blind spot.

For sentences edited by fixed rules (object swap, verb swap, action flip), zero lies within the confidence interval of
the difference between RA-RFT and the base (post hoc). The base already detects 77\% of object swaps.

On 222 sentences, the five Qwen-based grounders detect the action-flipped version of the sentence in the original video
less often (9.5--21.6 percentage points less; 22.9--40.0 on 35 rater-checked sentences, post hoc) than the original
sentence in the opposite-action video, contrary to the planned prediction. Detection rates are similar for sentences of
6 words or fewer (post hoc), so in longer sentences other details likely also conflict with the opposite-action video.

RaTSG, the second refusal-trained grounder that we were able to run, is trained to reject unrelated videos. With a 5\%
false-alarm rate set separately for the Charades development and test sets, RaTSG detects 28--41\% of unrelated
videos, and 8--10\% of opposite-action videos (post hoc).

\subsection{Reversed-frames check}
\label{sec:reversed-frames}

While the grounders' flags on the development set follow their forward yes/no answer (AUROC 0.66--0.77), they do not
follow the difference from frame reversal (0.48--0.52; post hoc). The direction information is therefore available but
not used. Grounding answers are also flagged when the MotionCD score (Section~\ref{sec:scoring}) exceeds a threshold
determined on the development set. On the 295 paired test sentences, the number of opposite-action detections increased
by applying frame reversal over the yes/no forward detection in all four Qwen settings (planned,
Table~\ref{tab:frame-check}; Qwen3-VL there uses the RA-RFT prompt, and for PerceptionLM the yes/no check alone adds
3.7 false alarms). For example, with the two-sentence prompt, applying this check increased opposite-action
detections from 39.7\% to 71.9\% in Qwen3-VL (post hoc, Figure~\ref{fig:main}c). The proportion of unrelated videos
detected on these sentences increased slightly, from 83.1\% to 84.4\%. With the check, the difference in detection between unrelated and
opposite-action videos decreased from 34.9, 38.3 and 30.8 to 4.7, 10.8 and 12.2 points (ArrowGEV, Time-R1 and RA-RFT
respectively; planned).

\begin{table}[!htb]
\centering
\small
\setlength{\tabcolsep}{3pt}
\begin{tabular*}{\columnwidth}{@{\extracolsep{\fill}}lrrrr@{}}
\toprule
 & False & \multicolumn{3}{c}{Opposite-action detections} \\
\cmidrule(l){3-5}
Grounder & alarms & Yes/no & Reversed & 95\% CI \\
\midrule
ArrowGEV     & 4.4 & $+12.5$ & $+33.9$ & $[+28.6,+39.2]$ \\
Time-R1      & 3.4 & $+11.2$ & $+29.2$ & $[+24.2,+34.4]$ \\
RA-RFT       & 3.7 & $+4.1$  & $+19.3$ & $[+15.0,+23.9]$ \\
Qwen3-VL (RA-RFT) & 5.1 & $+4.7$  & $+12.9$ & $[+9.1,+16.5]$ \\
PerceptionLM & 4.1 & $+24.4$ & $+35.9$ & $[+30.4,+41.6]$ \\
\bottomrule
\end{tabular*}
\par\medskip
\begin{tabular*}{\columnwidth}{@{\extracolsep{\fill}}lrrr@{}}
\toprule
 & \multicolumn{3}{c}{Opposite-action AUROC} \\
\cmidrule(l){2-4}
Reader & Yes/no & Shuffled & Reversed \\
\midrule
Qwen2.5-VL   & 0.764 & 0.757 & 0.839 \\
Qwen3-VL     & 0.747 & 0.764 & 0.842 \\
PerceptionLM & 0.883 & 0.855 & 0.926 \\
\bottomrule
\end{tabular*}
\caption{Opposite-action detection on 295 paired test sentences. Top: additional detections and false alarms in
percentage points, thresholds fixed on the development set; the interval is for the reversed-frames gain. Bottom: AUROC for opposite-action against original
videos, each score used alone.}
\label{tab:frame-check}
\end{table}

\begin{figure}[!htb]
\centering
\includegraphics[width=0.91\columnwidth]{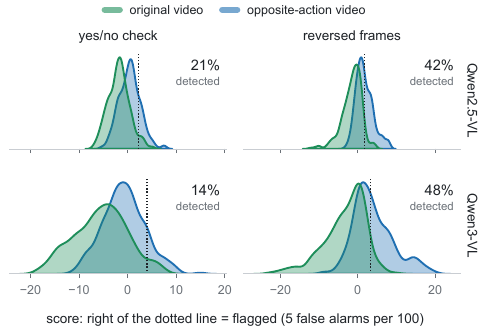}
\caption{Scores of the yes/no check and of reversed frames on original and opposite-action test videos. Dotted
line: threshold allowing 5 false alarms per 100 original videos; each panel gives the share detected (post hoc).}
\label{fig:scores}
\end{figure}

This is a specialist detection method. On its own, this score does not perform well in distinguishing unrelated from
original videos (AUROC = 0.68 and 0.67 for the yes/no answers of Qwen2.5-VL and Qwen3-VL, compared with 0.96 and 0.95
for the yes/no detection method; post hoc). Therefore, this detection method was added to the grounder's own flags.
The AUROCs in Table~\ref{tab:frame-check} are descriptive, except for PerceptionLM, where frame reversal gives a 0.044
higher AUROC than the yes/no check ($+0.019$ to $+0.070$; planned). Figure~\ref{fig:scores} shows the two scores on
original and opposite-action videos.

\begin{figure}[!htb]
\centering
\includegraphics[width=0.91\columnwidth]{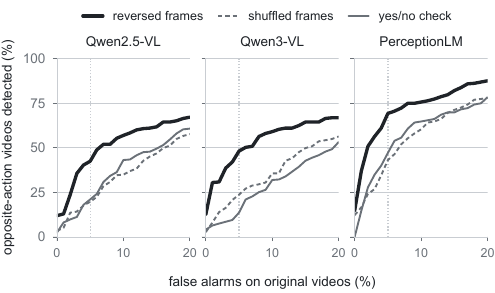}
\caption{Opposite-action videos detected against false alarms on original videos, for the yes/no check, shuffled
frames and reversed frames, each score used alone; one panel per model family, 295 paired test sentences,
thresholds set on the evaluated videos (post hoc).}
\label{fig:checks}
\end{figure}

In the shuffled condition, thresholds for both the reversed and the shuffled check were set on the original test videos so that no more than 5 additional false
alarms were made per 100 videos (Figure~\ref{fig:checks} shows every budget up to 20). The number of opposite-action detections increased by 10.8--18.6 points in Qwen2.5-VL
compared to shuffling (planned). For PerceptionLM, there was an increase of 17.6 points ($+8.0$ to $+29.1$; planned).
For Qwen3-VL using the RA-RFT prompt, the gain fell short of the planned criterion ($+3.4$ points [$-0.3$ to $+8.1$]).
With the two-sentence prompt, the increase was 14.6 points ($+6.5$ to $+20.7$; post hoc; adding shuffled frames gives 18.0
points and adding reversed frames 32.5).

Applied to the yes/no check, TAI~\cite{shin2026before} with its strength chosen on the development set gave an
increase in opposite-action detection of 8.8--24.1 points at a cost of 3.7--5.4 additional false alarms in 100 original
videos. The planned test that frame reversal outperforms TAI failed: it was ahead at 5 additional false alarms for 3
of the 4 grounders, but not at 10.

\section{Limits and Conclusion}
\label{sec:conclusion}

\paragraph{Limitations.}
Only results for household scenarios in Charades and Charades-Ego (recorded from the perspective of the subject wearing
the headset) are presented. For Charades, the absence of the queried event was verified by humans in 92 of 98
opposite-action videos. For Charades-Ego, humans found the queried event in 11 of the 38 opposite-action videos
reviewed, meaning its reported performance is derived from annotations alone. Annotations also missed the flipped
action of our rule-made edits in 5 of 40 original videos checked by a rater. Six video-language grounders from three
model families are tested, plus one grounder without a language component. One refusal-trained method of each kind is
used. Each model has been run once with greedy decoding. Re-running inference with the same batches gives identical
results; however, altering the batches changes 2--3\% of flags. As Charades-STA sentences are commonly used in
training, and it is unclear how much exposure models have had to these sentences, results are only compared on
identical items. While new queries have been used in testing, the test set reuses videos from the development set.

\paragraph{Conclusion.}
Just as in anomaly detection, it is most challenging to detect an absent event in a video when it contains a similar
event. Benchmarks using unrelated videos or altered queries do not capture this situation. Training on altered queries
provides a benefit mainly for those examples. Therefore, evaluations should include real videos containing the
opposite action to the query. A training-free check can detect many missed opposite-action videos by comparing
responses to forward and reversed frames. This adds few false alarms, and can be used as a baseline for future
methods.

\section*{Acknowledgments}
We thank our good friend Ruman Fatma for watching and rating videos in the label audit (Section~\ref{sec:viewing}).

\FloatBarrier

\end{document}